\documentclass{article} 
\usepackage{nips15submit_e,times}
\usepackage{hyperref}
\usepackage{url}

\title{Can Pretrained Neural Networks Detect Anatomy?}

\author{
Vlado Menkovski \\
Philips Research\\
Eindhoven, Netherlands \\
\texttt{vlado.menkovski@philips.com} \\
\And
Zharko Aleksovski \\
Philips Research \\
Eindhoven, Netherlands \\
\texttt{zharko.aleksovski@philips.com} \\
\AND
Axel Saalbach \\
Philips Research \\
Hamburg, Germany \\
\texttt{axel.saalbach@philips.coom} \\
\And
Hannes Nickisch\\
Philips Research \\
Hamburg, Germany \\
\texttt{hannes.nickisch@philips.com} \\
}

\nipsfinalcopy 

\begin{document}

\maketitle

\begin{abstract}
Convolutional neural networks demonstrated outstanding empirical results in computer vision and speech recognition tasks where labeled training data is abundant. In medical imaging, there is a huge variety of possible imaging modalities and contrasts, where annotated data is usually very scarce. We present two approaches to deal with this challenge. A network pretrained in a different domain with abundant data is used as a feature extractor, while a subsequent classifier is trained on a small target dataset; and a deep architecture trained with heavy augmentation and equipped with sophisticated regularization methods. We test the approaches on a corpus of X-ray images to design an anatomy detection system. 
\end{abstract}

\section{Introduction and Motivation}

Deep learning \cite{lecunn15DL} (DL) is rapidly gaining momentum in computer vision \cite{krizhevsky12DCNN}, speech recognition \cite{hinton12deep} and medical imaging applications due to their record-breaking empirical performance and their promise to jointly learn the low-level features and the high-level decisions directly from raw data.

One major open research challenge is the transfer of the excellent predictive performance of DL methods, in particular convolutional neural networks (CNN),
from the ``data-laden'' regime of computer vision\footnote{There are $10^{6}$ images in $10^{3}$ classes in the ILSVRC challenge, \url{http://www.image-net.org/challenges/LSVRC}.} to the ``data-bounded'' regime of medical imaging.

Since the typical number of parameters in the DL models is very big the major concern for operating in ``data-bounded'' regimes is inability to generalize and form robust features because the small number of examples do not capture well the variability in the input space. There are various approaches to deal with this challenge. Data augmentation can be used to increase the effective size of the training set and drive the training to build invariances to specific transformations. Network architecture design can be used to hard-code prior knowledge (such as translation invariance) by design. Sophisticated regularization methods such as dropout \cite{srivastava2014dropout} and batch normalization (BN) \cite{ioffe2015batch} can work to counter overfitting. Finally, \emph{global} training of the feature extraction layers on abundant data domains (i.e. natural images) and then training the ``classification'' higher layers from the limited data domains can be used to ensure that the lower-level features are invariant to the particularities of the training dataset.

We choose two directions to investigate these approaches for the goal of anatomy detection in X-ray images. The first using a network trained on a large corpus of natural images for feature extraction combined with a SVM \cite{cortes1995support} classifier trained on the medical images. And a second approach using heavy augmentation and regularization of a deep neural network trained only on the small-sized medical image dataset. 

\section{Experiments and Results}

In order to assess the performance of specifically trained and off-the-shelf networks, radiographs from the ImageClef 2009 - Medical Image Annotation task\footnote{\url{http://www.imageclef.org/2009/medanno}} were used. This challenge database consists of a broad range of X-ray images from clinical routine, with a detailed anatomical classification.
For convenience, we flattened the hierarchical class representation and disregarded classes with fewer than 50 example images. In doing so we ended up with 24 unique classes. For evaluation purposes, the entire data corpus (of 14676 images) was divided into a training and test set, covering 90\% and 10\% of the data respectively.

As a first baseline, the OverFeat network (see \cite{sermanet14overfeat}) in combination with a (linear) multi-class SVM was employed. OverFeat is a convolutional neural network that was trained on 1.2 million non-medical images form the ImageNet2012 database with 1000 classes, and has been successfully employed as a generic feature extractor in different applications.

For the purpose of this experiment, we selected the \emph{fast} OverFeat instantiation and used the default setting for feature extraction. This results in a 4096 dimensional representation of the images, which was subsequently used as input for the SVM classier. The regularization parameter C of the linear SVM was estimated on the training set using exhaustive search and 5-fold cross-validation. 

For the second direction we designed a network \ref{table:arch} inspired by the work of Razavian et al. \cite{sharif2015baseline}, using dropout and batch normalization with leaky ReLU activation functions \cite{he2015delving}. In this case for training we used images of 128 by 128 pixels with various augmentations ranging from resizing and cropping, rotation, translation, shearing, stretching and flipping. The results of the two directions are given in table \ref{table:results}.

\begin{table}[ht]
\caption{Results: Accuracy in percent}
\centering 
\begin{tabular}{c c}
\hline\hline
OverFeat + SVM & Custom CNN \\ [0.5ex] 
\hline 
92.42 & 95.14 \\ [1ex] 
\hline 
\end{tabular}
\label{table:results}
\end{table}

\section{Conclusions and Perspectives}

We investigated when, where, and how to (re)train a convolutional neural network used for anatomy detection. Our results suggest that pretrained networks are good image descriptors even outside their training domain and that retraining only the last layer is a viable alternative to perform transfer learning in light of
limited training data. We also demonstrated that sufficiently expressive models such as a very deep neural network can be trained even on relatively small number of annotations if proper augmentation and regularization is implemented. Our work is a first step towards a multi-modality anatomical expert system with components trained to maximize data efficiency.

\bibliographystyle{abbrv}
\bibliography{deep}

\section*{Annex}
\begin{table}[ht]
\caption{Custom Deep Neural Network architecture}
\centering 
\begin{tabular}{c c}
\hline\hline
ConvLayer + BN       &       (3x3, 32x) \\
ConvLayer + BN       &      (3x3, 16x) \\ 
MaxPoolLayer         & 		(3x3, 2x2 stride) \\ 
ConvLayer + BN       &      (3x3, 64x) \\ 
ConvLayer + BN       &      (3x3, 32x) \\ 
MaxPoolLayer         & 		(3x3, 2x2 stride) \\ 
ConvLayer + BN       &      (3x3, 128) \\ 
ConvLayer + BN       &      (3x3, 128) \\ 
ConvLayer + BN       &      (3x3, 64) \\ 
MaxPoolLayer         & 		(3x3, 2x2 stride) \\ 
ConvLayer + BN       &      (3x3, 256) \\ 
ConvLayer + BN       &      (3x3, 256) \\ 
ConvLayer + BN       &      (3x3, 128) \\ 
MaxPoolLayer + Dropout 	&	(3x3, 2x2 stride) \\ 
DenseLayer + BN + Dropout & (256) \\ 
DenseLayer + BN + Dropout & (256) \\ 
Softmax		               & (24) \\ 
\hline 
\end{tabular}
\label{table:arch}
\end{table}

\end{document}